\documentclass[10pt,twocolumn,letterpaper]{article}

\usepackage{iccv}
\usepackage{times}
\usepackage{epsfig}
\usepackage{graphicx}
\usepackage{amsmath}
\usepackage{amssymb}
\usepackage{booktabs}
\usepackage{multirow}
\usepackage{subfig}

\usepackage[pagebackref=true,breaklinks=true,letterpaper=true,colorlinks,bookmarks=false]{hyperref}

\iccvfinalcopy 

\def\iccvPaperID{3991} 

\ificcvfinal\fi
\begin{document}

\title{Instance Scale Normalization for image understanding}

\author{Zewen He$^{1}$\thanks{Equal contribution.}
~~~~~He Huang$^{1}$\footnotemark[1]
~~~~~Yudong Wu$^{1}$\footnotemark[1]
~~~~Guan Huang$^{2}$
~~~~~Wensheng Zhang$^{1}$\\
$^1$Institute of Automation, Chinese Academy of Sciences, Beijing, China\\
$^2$Horizon Robotics, Beijing, China\\
{\tt\scriptsize \{hezewen2014, huanghe2016, wuyudong2016\}@ia.ac.cn,   zhangwenshengia@hotmail.com,   guanhuang@horizon.ai}}


\maketitle

\begin{abstract}
   Scale variation remains a challenging problem for object detection.
   Common paradigms usually adopt multi-scale training \& testing (image pyramid) or FPN (feature pyramid network)
   to process objects in a wide scale range.
   However, multi-scale methods aggravate more variations of scale that even deep convolution neural networks with FPN cannot handle well.
   In this work, we propose an innovative paradigm called Instance Scale Normalization (ISN) to resolve the above problem.
   ISN compresses the scale space of objects into a consistent range (ISN range), in both training and testing phases.
   This reassures the problem of scale variation fundamentally and reduces the difficulty of network optimization.
   Experiments show that ISN surpasses multi-scale counterpart significantly for object detection, instance segmentation, and multi-task human pose estimation, on several architectures.
   On \textit{COCO} test-dev, our single model based on ISN achieves 46.5 mAP with a ResNet-101 backbone,
   which is among the state-of-the-art (SOTA) candidates for object detection.
\end{abstract}

\section{Introduction}
The vision community has rapidly improved performance on object recognition,
especially object detection~\cite{ren2015faster}, instance segmentation~\cite{he2017mask},
human pose estimation~\cite{he2017mask}.
Among them, Object detection is a prerequisite for many downstream applications.
The performance of detectors has dramatically made progress
with the help of powerful backbone networks~\cite{he2016deep}, delicate design on optimization objective~\cite{lin2017focal}
and well-annotated datasets~\cite{lin2014microsoft}.

However, detecting objects of various scales remains challenging, especially encountering objects of extreme size.
As shown in Table ~\ref{intro_ap_on_scales}, AP on small objects falls much compared to medium and large objects.
To alleviate the scale variation problem, state-of-the-art detectors rely on feature pyramids~\cite{lin2017feature} or image pyramids\cite{felzenszwalb2010object}.
On the one hand of feature pyramids, FPN~\cite{lin2017feature} construct a multi-stage network with the parallel prediction on objects of isolated scale range.
On the other hand of image pyramids, simple multi-scale training \& testing strategy still play a role in multiple recognition tasks.
In particular, ~\cite{singh2018analysis, singh2018sniper, najibi2018autofocus} found that ignoring loss signals from 
extremely tiny and large objects can improve detection accuracy.

It should be pointed out that the aforementioned methods have defects respectively. 
According to ~\cite{wang2019region}, the receptive field and semantic scope of the same RoI should be consistent in the feature map.
For the object of large scale, the receptive field may not be sufficient; while for the small one, the semantic scope is larger comparing to object's size.
So there exists consistency only when object's size falls into a moderate range.
Multi-scale training resizes images to different resolutions, therefore resizing some objects to normal scales.
But it also lets some objects to extreme scales with great inconsistency, which gives rise to final accuracy degradation.
As shown in Table ~\ref{intro_mstrain_ap}, the detector trained from image pyramid with a wider scale range is inferior to the one from the normal scale range.
FPN ~\cite{lin2017feature} employed a heuristic rule to make feature maps on different stages responsible for RoIs of isolated scale range.
It also suffers from the inconsistency when encountering tiny objects (downsampled to sub-pixels and hard to recognize) or large objects (receptive field cannot cover).
SNIP ~\cite{singh2018sniper} tried to ignoring the extremely tiny and large objects, resulting in removing the inconsistency on these extreme samples.
But there also exists inconsistency about the scale range usage. 
The detailed analysis will be reported in ~\ref{ISN_range_calc}.

We propose a simple and effective method approach, called Instance Scale Normalization (ISN), to further alleviate the inconsistency resulting from scale variation.
ISN will also train detector on image pyramid, but only optimize model on objects in moderate scale range.
With FPN integrated, the heuristic rule in FPN can distribute RoIs of large scale range to multiple stages which process RoIs in smaller range.
This alleviates the learning difficulty and enlarges feasible range on object scales, resulting in more available training samples and better generalization.
Based on ISN, our contribution can be summarized as follows:
\begin{itemize}
\item We propose ISN to restricting RoIs of extreme scale in training and testing, this strategy can alleviate the negative impact caused by large scale variation.
Without any modification to network structure,
SOTA result of the single ResNet-101 model is obtained by ISN on the COCO object detection benchmark.
\item We extend ISN to other recognition tasks, i.e., instance segmentation and keypoint detection.
Comparable
improvements have been achieved, especially for keypoint
detection (+3.6\% AP), which is more sensitive
to scale variation.
To the best of our knowledge, best performance of a single ResNet-50 model can be achieved on both tasks by ISN.
\item ISN benefits detectors of various backbones, especially tiny ones like ResNet-18 and MobileNet-v2 (e.g., 
On ResNet-18 based Faster R-CNN, ISN boost mAP by 5 point than original multi-scale training \& testing).
For fast inference purpose, 
the models are usually tested on a single scale.
We found model trained with ISN still achieves better performance
 than the ordinary one in this situation.
This makes ISN a cost-free freebie for object recognition task
in practical applications.
\end{itemize}

\begin{table}
   \small
   \centering
   \begin{tabular}{c|cccccc}
      \toprule
         FPN &AP  &AP$_S$  &AP$_M$  &AP$_L$  \\
      \midrule
            &34.7 &16.3    &38.2    &49.3 \\
$\checkmark$&38.0 &22.0    &41.5    &48.8 \\
      \bottomrule
   \end{tabular}
   \vspace{0.1cm}
   \caption{\small{Results of different scales on Faster R-CNN of ResNet-50}}
   \label{intro_ap_on_scales}
\end{table}

\begin{table}
   \small
   \centering
   \begin{tabular}{c|cccccc}
      \toprule
         image pyramids             &AP   &AP$_S$  &AP$_M$  &AP$_L$  \\
      \midrule
         img-scale [640, 800]   &28.8 &13.0    &30.8    &40.8    \\
         img-scale [160, 1600]  &27.8 &12.3    &29.4    &40.4    \\
      \bottomrule
   \end{tabular}
   \vspace{0.1cm}
   \caption{\small{
   We train two detectors using image scales randomly from a normal range $[640, 800]$ 
   and larger range [160, 1600] respectively. Both detectors are tested on single scale $800$. The accuracy
   from larger scale range is worse than the normal one. Detectors are simple Faster R-CNN on ResNet-18.}}
   \label{intro_mstrain_ap}
\end{table}

\section{Related Works}
   Driven by representation capacity of deep conv-feature, CNN-based detectors~\cite{girshick2014rich, ren2015faster, liu2016ssd, redmon2016you}
are a dominant paradigm in object detection community. The R-CNN series and variants~\cite{ren2015faster, girshick2015fast, ren2015faster}
gradually promotes the upper bound of performance on two-stage detectors. In particular, Faster R-CNN~\cite{ren2015faster} adopts a shared
backbone network to proposal generation and RoI classification, resulting in real-time detection and accuracy rising.
Mask R-CNN~\cite{he2017mask} introduced a multitask framework for related recognition tasks, such as instance segmentation and human pose estimation,
achieving higher accuracy with brief procedure and structure. On the other hand, YOLO-v3~\cite{redmon2018yolov3}, RetinaNet~\cite{lin2017focal}, CornerNet~\cite{law2018cornernet}
and etc, rely on only one stage to predict while struggling to catch up with the top two-stage detectors.
To compare and verify effectiveness of the proposed method, ISN model is implemented based on Faster R-CNN~\cite{ren2015faster} and Mask R-CNN~\cite{he2017mask} for various tasks.

There still exist difficulties for detectors on objects with large scale variation. Early solutions choose to learn scale-invariant representation for objects to tackle scale variation problems.
For traditional members, Haar face detector~\cite{viola2001rapid, viola2004robust} and DPM~\cite{felzenszwalb2010object} become more scale-robust with the help of image pyramid~\cite{adelson1984pyramid}.
SSD~\cite{liu2016ssd}, SSH~\cite{najibi2017ssh}, MS-CNN~\cite{cai2016unified} try to detect small objects at lower layers, while big objects at higher layers.
To be further, FPN~\cite{lin2017feature} fuses feature maps at adjacent scale to combine semantics from upper and details from lower.
Objects of different size can be predicted at corresponding levels according to a heuristic rule. SAFD\cite{hao2017scale} predicts scale distribution histogram of face,
which guides zoom-in and zoom-out for face detection. FSAF\cite{zhu2019feature} selected the most suitable level of feature for each object dynamically in training compared with FPN.
SNIP~\cite{singh2018analysis} assumes it's easy for detectors to generalize well to objects of moderate size. Only objects of normal scale range are utilized to make model converge better.
Ulteriorly, SNIPER\cite{singh2018sniper} effectively mined on generated chips for better result.

Compared to SNIP etc~\cite{singh2018analysis, singh2018sniper}, ISN adopts an instance scale normalization strategy to select training samples
and integrates FPN to attain better generalization. It also extends to other recognition tasks, validating the effectiveness of our method.

\section{Method}

Current models still suffer from large scale variation and 
cannot obtain satisfactory accuracy even with multi-scale
training \& testing.
We will introduce our \textit{instance scale normalizatoin}(ISN) in this section to deal with it better.
Concretely, in section \ref{frcnnrecap}, the common Faster R-CNN detector is recapped.
In section \ref{ISN_range_calc}, we analyze the drawbacks of SNIP which motivates the proposal of ISN.
In section \ref{method_ISN_range}, we detail the object sampling mechanism of ISN technique, including
the effects on FPN.
In section \ref{isnonrecognition}, details of ISN on various recognition tasks are described.

\subsection{Faster R-CNN detector recap}\label{frcnnrecap}
Faster R-CNN and its variants are leading detectors in object detection community, currently.
They basically consist of two steps.
In the 1st-stage, region proposal network (RPN) generates a bunch of RoIs (Region of Interest) on basis of a set of pre-defined anchors.
These RoIs indicate the region where possible objects exist in the image.
Then, in the 2nd-stage, a Fast RCNN\cite{girshick2015fast} 
extracts fixed-size feature (e.g., $7\times7$) for each RoI from the shared feature with RPN.
This can be implemented by RoIPooling~\cite{ren2015faster} or RoIAlign~\cite{he2017mask} operators.
Finally, these features will be sent to two independent subnets for category classification and box regression, respectively.

In Faster R-CNN, all ground-truth object bounding-boxes (gt-bboxes) in current image are collected to participate in training.

\subsection{Object scale range}\label{ISN_range_calc}
We think that ConvNets intrinsically suffer from large scale variation in training, as shown in Table~\ref{intro_ap_on_scales}.
An ordinary solution is diminishing the large variation, namely sampling objects in moderate scale range.
SNIP~\cite{singh2018analysis} gives a detailed analysis on the effect of object-scale to training.
However, we argue that there exists inconsistency in SNIP's object selection mechanism.

\begin{figure}[h]
   \subfloat[Original (480,640)]{
      \begin{minipage}[t]{0.48\linewidth}
         \centering
         \includegraphics[width=4cm]{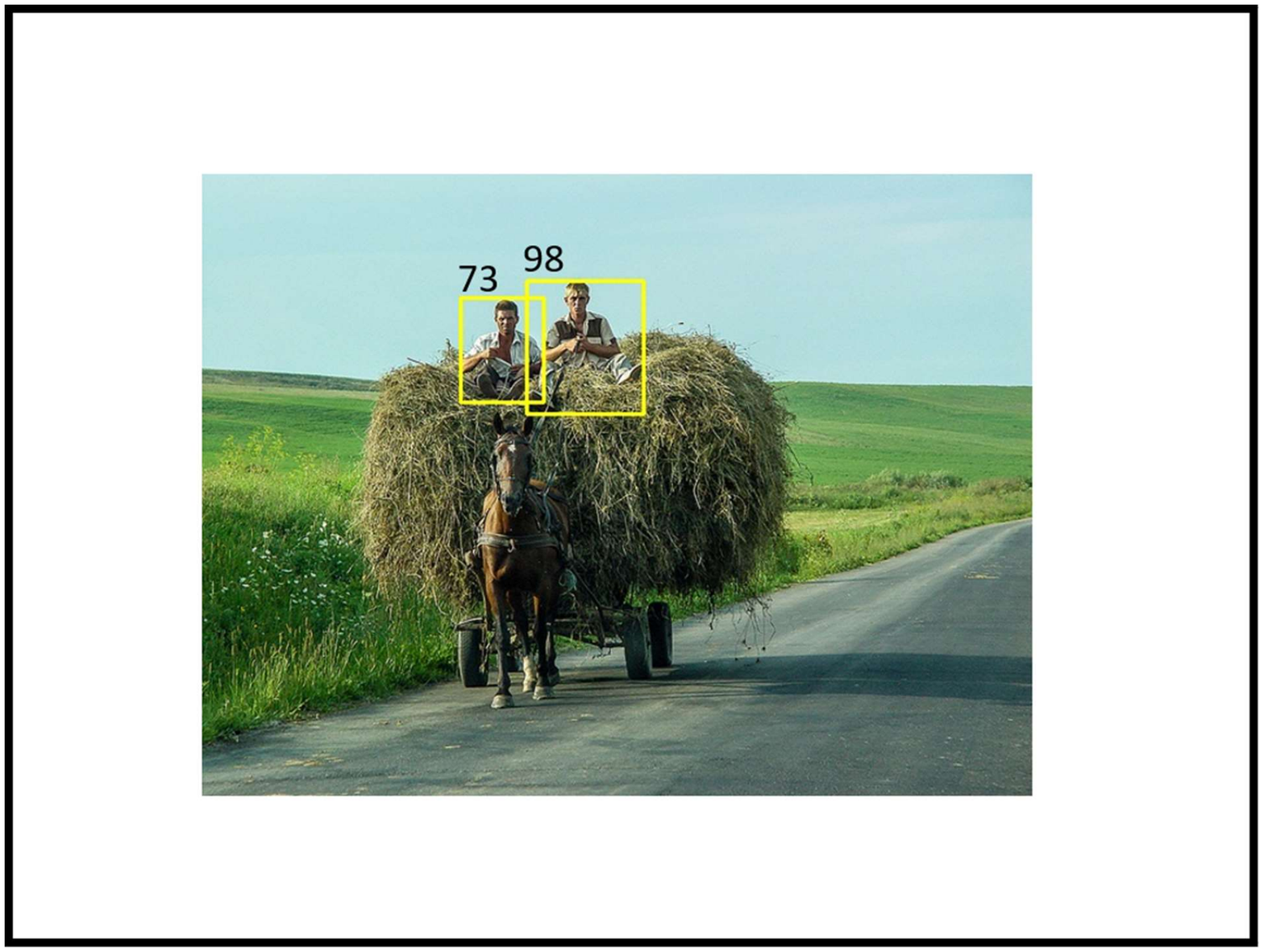}
      \end{minipage}
      \label{fig:SNIP_ISN_compare_ori1}
   }
   \subfloat[SNIP (800,1200)]{
      \begin{minipage}[t]{0.48\linewidth}
         \centering
         \includegraphics[width=4cm]{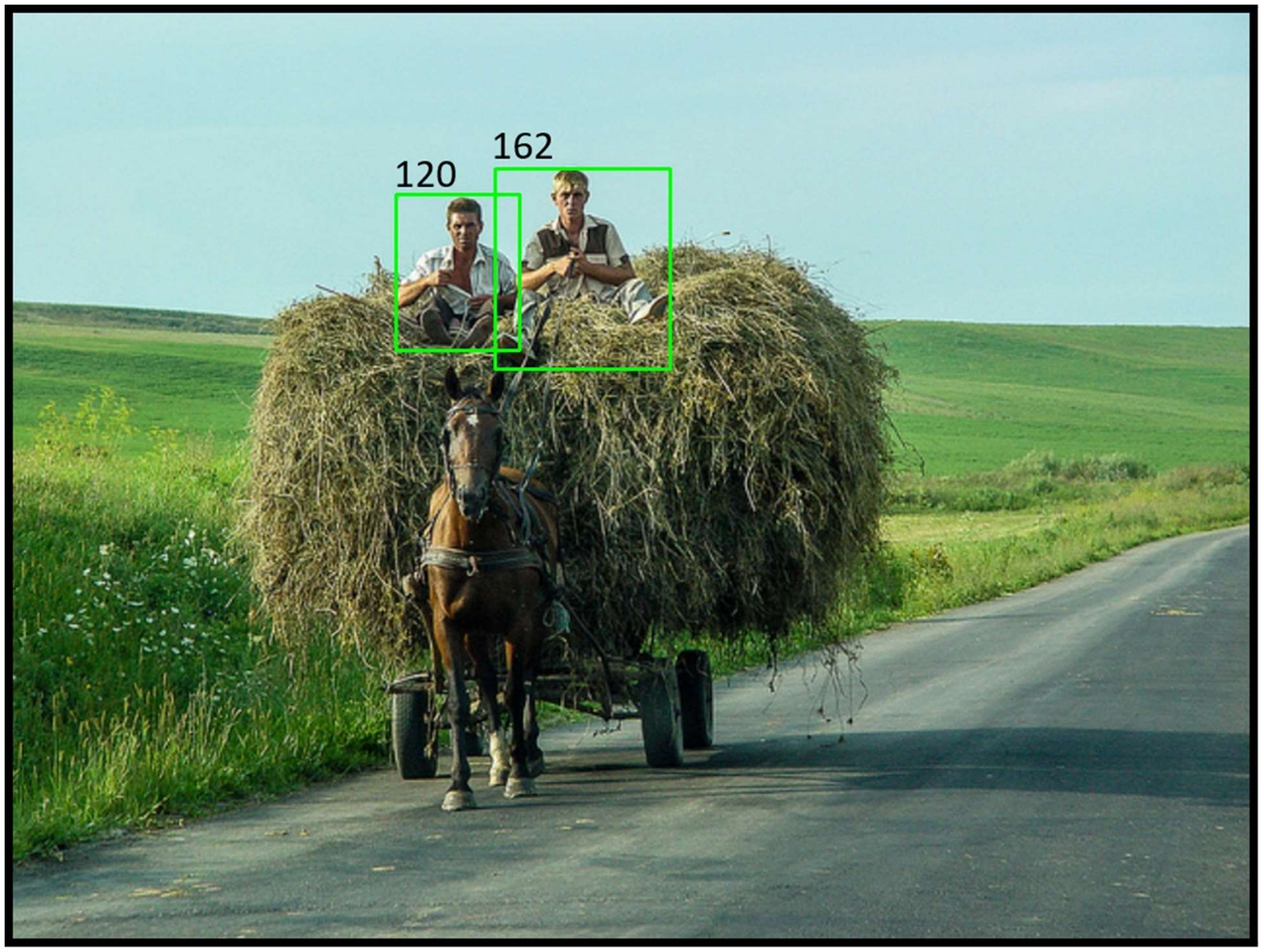}
      \end{minipage}
      \label{fig:SNIP_ISN_compare_snip1}
   }
   
   \subfloat[Original (425,640)]{
      \begin{minipage}[t]{0.48\linewidth}
         \centering
         \includegraphics[width=4cm]{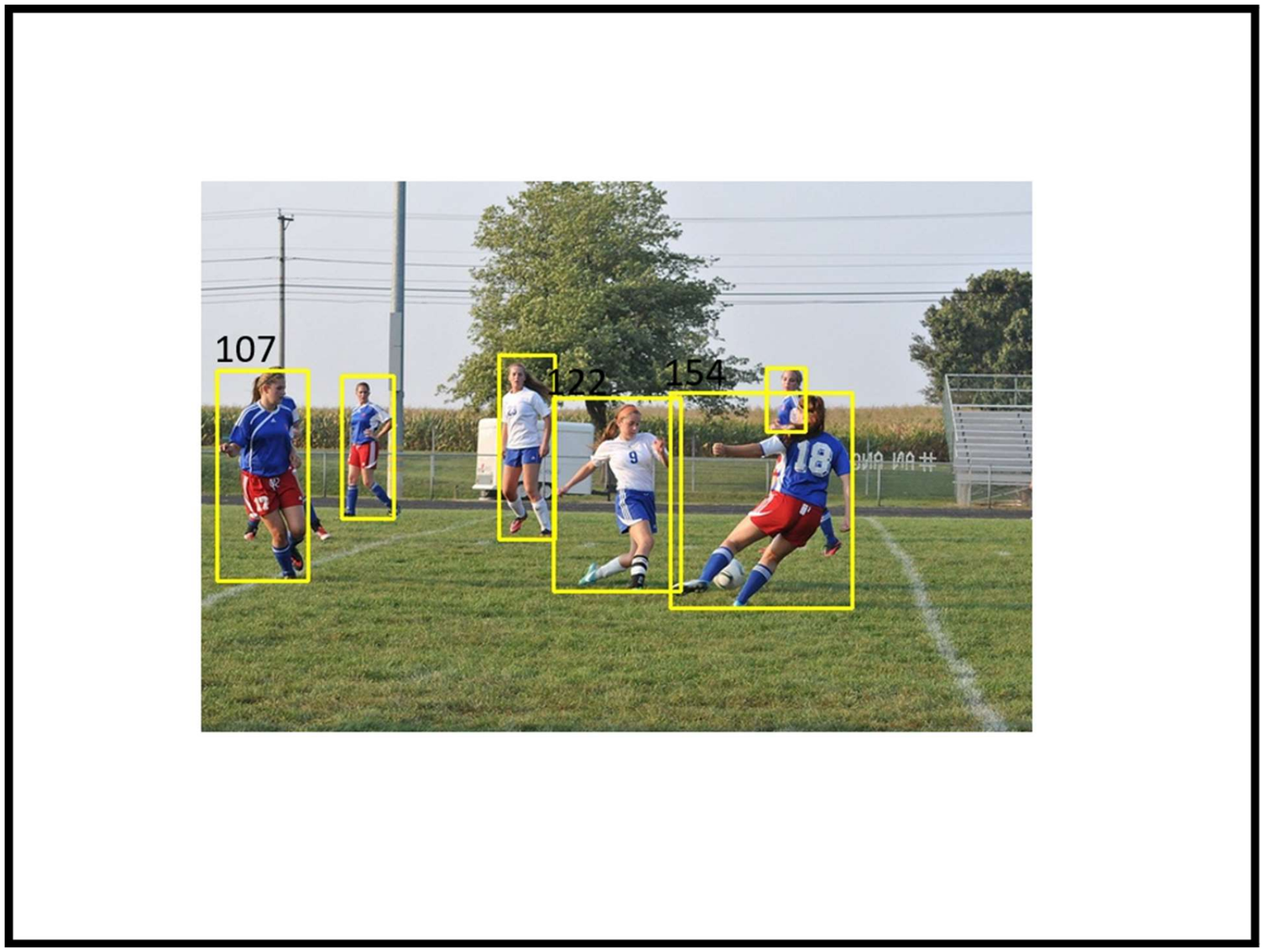}
      \end{minipage}
      \label{fig:SNIP_ISN_compare_ori2}
   }
   \subfloat[SNIP (480,723)]{
      \begin{minipage}[t]{0.48\linewidth}
         \centering
         \includegraphics[width=4cm]{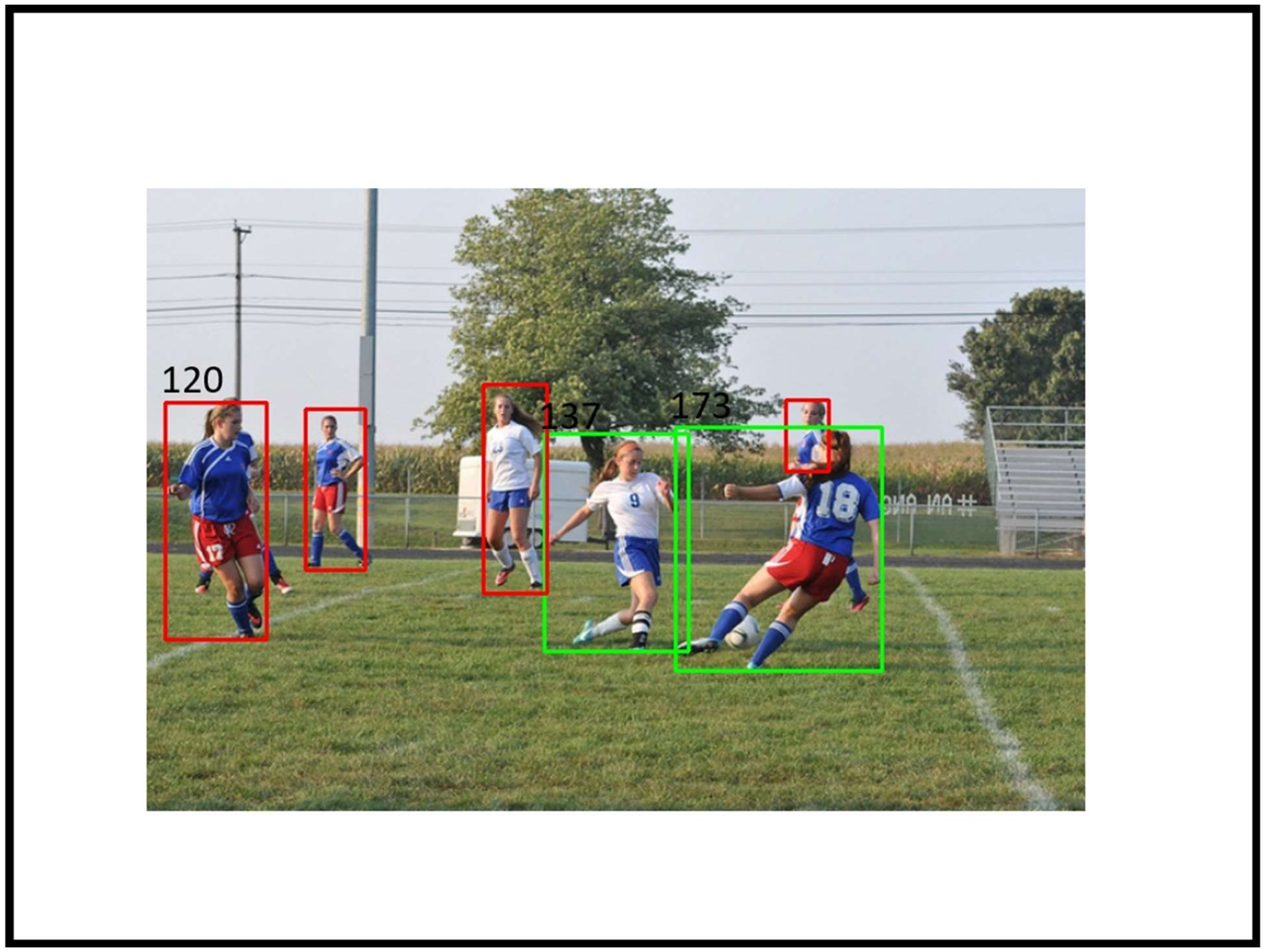}
      \end{minipage}
      \label{fig:SNIP_ISN_compare_snip2}
   }
   \vspace{-0.3cm}
   \subfloat[Objects trained in SNIP]{
      \begin{minipage}[t]{0.48\linewidth}
         \centering
         \includegraphics[width=4cm]{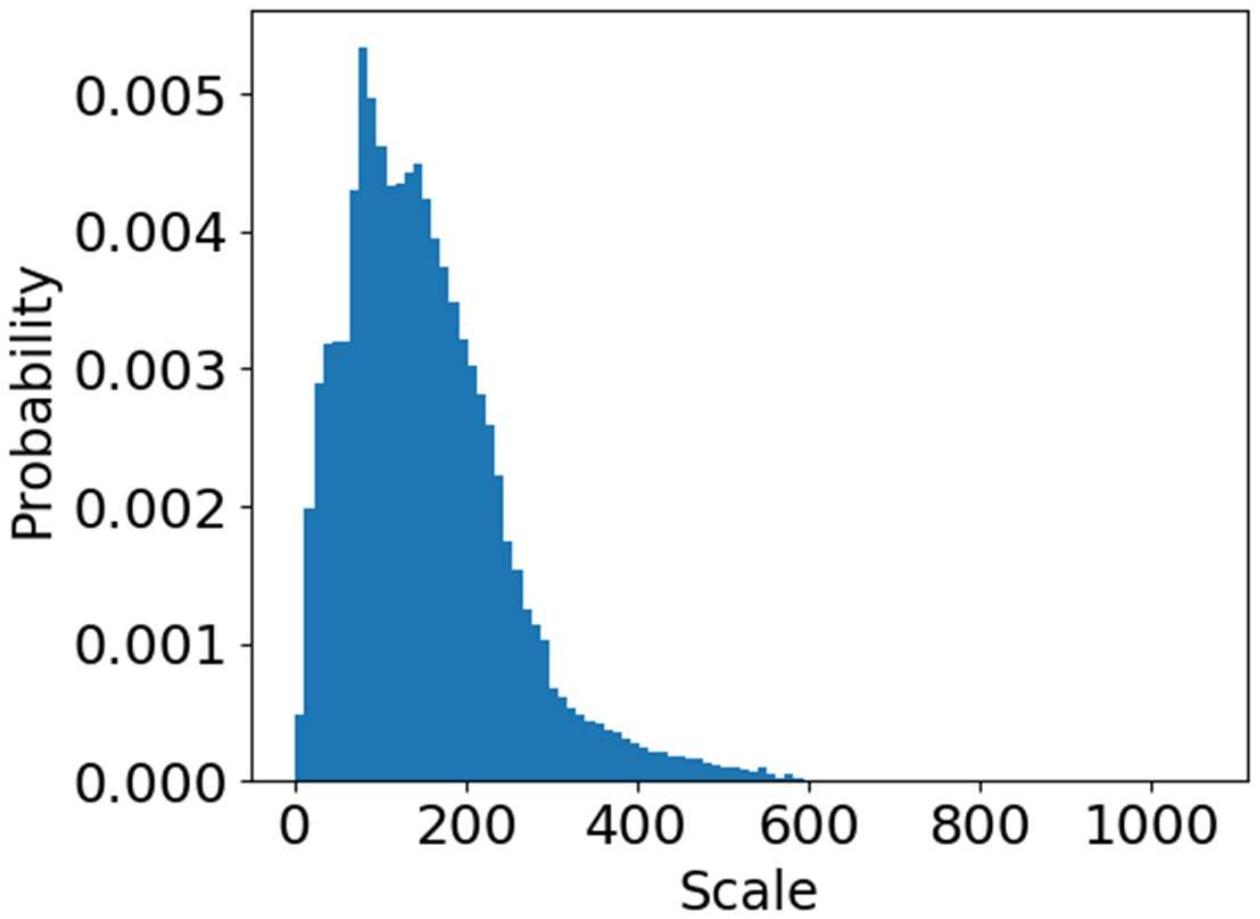}
      \end{minipage}
      \label{fig:SNIP_ISN_compare_objectsused}
   }
   \subfloat[Objects ignored in SNIP]{
      \begin{minipage}[t]{0.48\linewidth}
         \centering
         \includegraphics[width=4cm]{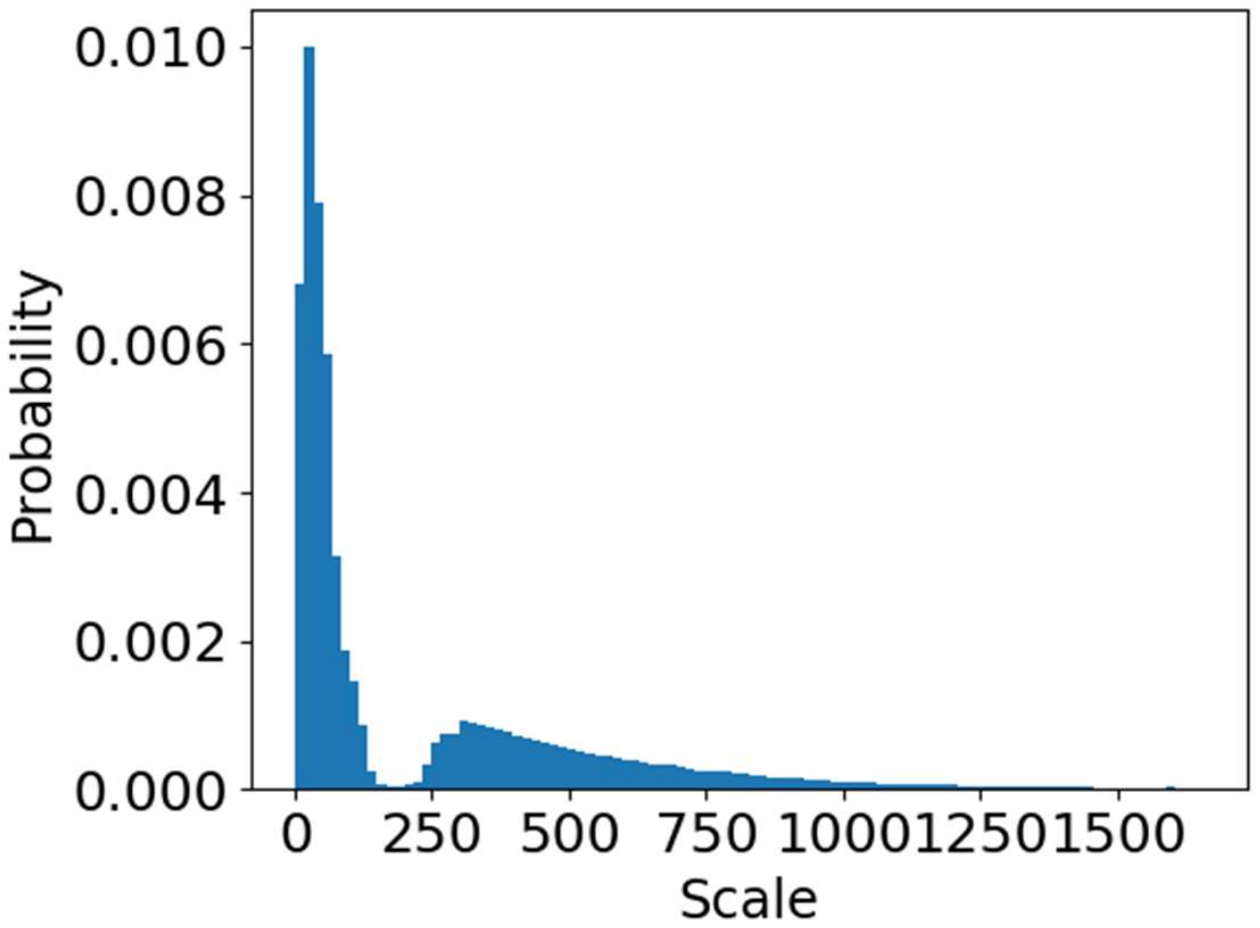}
      \end{minipage}
      \label{fig:SNIP_ISN_compare_objectsignored}
   }
   \caption{\small{
      \textbf{Problem from inconsistency in SNIP}:
      For (a)-(d), each bbox with red edge means objects ignored in SNIP, and green bbox means used by SNIP.
      The black number at top right of each bbox means the corresponding scale in that image.
      In SNIP~\cite{singh2018analysis}, valid range in the original image
      for resolution $(800, 1200)$, $(480, 800)$ are $(40, 160)$ and $(120, \inf)$.
      (a) and (c) are resized to (b) and (d).
      The man at left in image (b) will be retained because its scale is $73\in(40,160)$;
      while the woman player at the leftmost in image (d) will be ignored because its scale is $107\notin(120,\inf)$.
      When these two have the same scales, 
      they are regarded as different roles in training.
      For (e)-(f), the former shows the scale distribution of objects in resized images used in SNIP training,
      while the latter shows the distribution of ignored ones.
      There exists a big overlap between them.
   }}
   \label{fig:SNIP_ISN_compare}
\end{figure}

\subsubsection{Inconsistency in SNIP}\label{sniproblem}
SNIP is a training scheme designed for multi-scale training, e.g., with 
image pyramid.
It aims at excluding objects of extreme size in the training phase.
In detail, the valid training range in original image is carefully tuned for each image resolution.
Then for the $i$-th resolution, 
the RoIs whose area (w.r.t original image) fall in the valid range $[s_i^c, e_i^c]$
will participate in training. Otherwise, they are ignored.


However, there is a basically unreasonable case with this training paradigm.
That is, the objects with nearly the same scale in resized images may not take part in training together. 
As illustrated in Fig~\ref{fig:SNIP_ISN_compare},
the man on the left in Fig~\ref{fig:SNIP_ISN_compare}(b)
and woman player on the leftmost in Fig~\ref{fig:SNIP_ISN_compare}(d) shares the same scale in resized resolutions.
The former can take part in training while the latter cannot.

Fig~\ref{fig:SNIP_ISN_compare}(e) gives the distribution of
training objects in SNIP.
As it shows, there are plentiful extremely tiny, and huge objects participating in the training phase.
The Fig~\ref{fig:SNIP_ISN_compare}(f) exhibits the distribution of
ignored objects in SNIP. 
As it is shown, 
the ignored objects overlap much with the trained objects, 
which is the cause of the contradiction in previous unreasonable case.
In a word, the ignoring mechanism is implicit and inconsistent in SNIP.
So, the model's behavior is uncertain and the most suitable scale of the model for testing is indeterminate.
In SNIP~\cite{singh2018analysis}, the valid range for testing is obtained via a greedy search in val-set.

So we argue the valid range here is not consistent between training 
and testing phase, as well as among different resolutions.
This results in burdensome tuning of hyper-parameters.
For example, for a three-level image pyramid,
one needs to tune three pairs of valid ranges for training and another three pairs for testing.
The inconsistency could also hamper the model performance due to self-contradictory sampling strategy.

\begin{figure*}
   \centering
   \includegraphics[clip, trim=4cm 0cm 6cm 0cm, width=14cm, height=5cm]{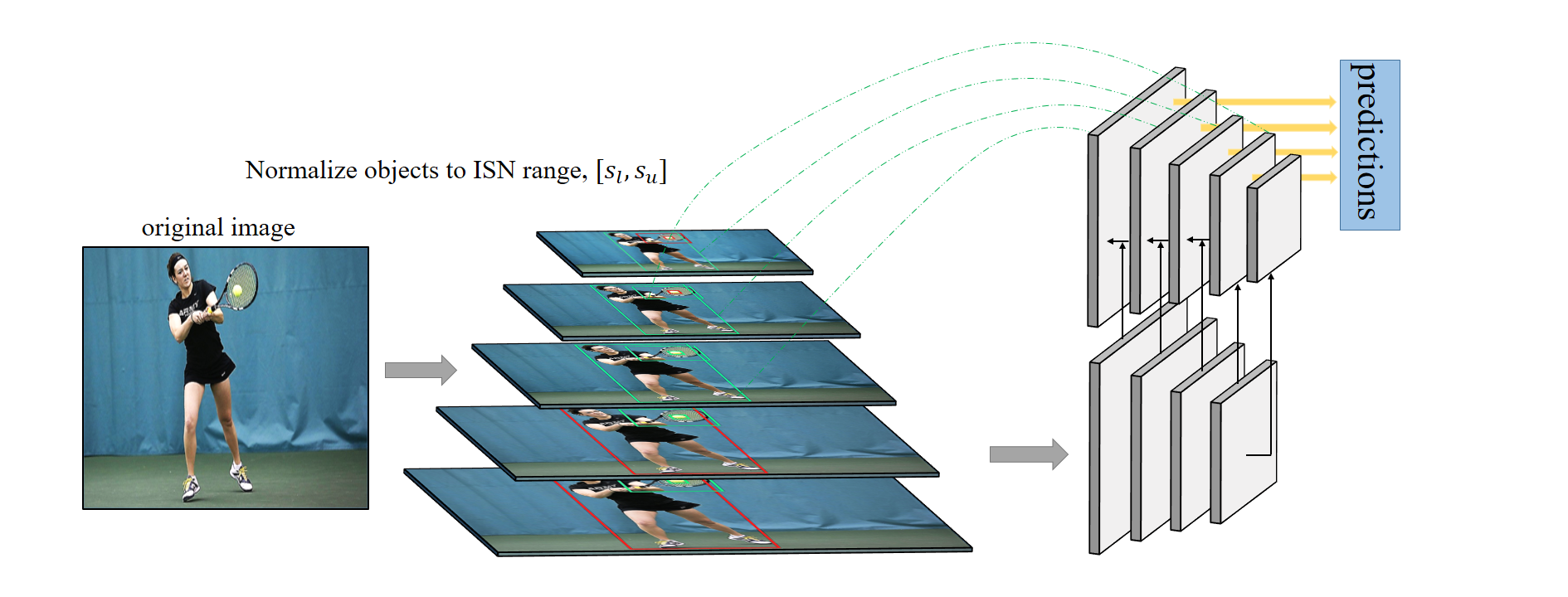}
   \caption{\small{\textbf{ISN plus FPN}:
   Here shows ISN for multi-scale training \& testing.
   In training, ISN firstly resizes original image $\omega_i$ times to get $i$-th resolution.
   Then it select objects which scale falls in ISN range (marked by green boxes) as valid for training,
   otherwise discards invalid boxes (red).
   In testing, only the predicted boxes from different resolutions in ISN range are preserved and fused.
   }}
   \label{fig:ISN_FPN}
\end{figure*}

\vspace{-0.05cm}

\subsection{Instance Scale Normalization}\label{method_ISN_range}
Previous scale normalization method, e.g., SNIP has
some self-contradiction in its sampling strategy.
Here, ~\textit{instance scale normalization} (ISN) method is an improved multi-scale training \& testing strategy
with instance scale normalization on objects.
\subsubsection{Instance scale normalization}
With the same chip generation strategy in SNIPER~\cite{singh2018sniper},
ISN employs the scale normalization that adopts a consistent scale range $[s_l, s_u]$, called ISN range, for different phases and resolutions.
In the training phase, firstly, scaling factor set $\Omega$ is pre-defined.
For the original image with size $[h, w]$, 
it is resized to $i$-th resolution $[h*\omega_i, w*\omega_i]$ by $\omega_i\in\Omega$.
Then each RoI whose scale on $i$-th resolution falls in this range will be set to valid currently, else invalid, as shown in Fig~\ref{fig:ISN_FPN}.
The invalid objects constitute the ignored regions that don't contribute to training.
In the testing phase for each resized resolution, all predicted RoIs whose scale falls within the same ISN range will be kept for post-processing.
In this way, all objects can be normalized to a moderate scale range consistently.
This improved selection strategy is obviously more excellent.
On the one hand, the uniform scale range eliminates the adaption of scale between training and testing phase.
On the other hand, the instance scale normalization reduces scale fluctuation of objects trained on different image resolutions.
What's more, ISN only tunes the two parameters, i.e., ISN range $[s_l, s_u]$, no matter how many image resolutions are used,
while SNIP is in linear with that.
For example, for a 3-level image pyramid, the number of scale range parameters is 12 with SNIP.
\subsubsection{Feature pyramid integration}\label{isnonfpn}
When setting scale range limitation for objects, the number of valid objects will decrease, resulting in possible over-fitting, as ~\cite{li2019scale} shows.
Even taking multi-scale training method, this over-fitting influence cannot be eased.

FPN adopts separate stages, i.e., $P_2, P_3, P_4, P_5$, to predict objects at disjoint scale range,
and has better capability for detecting objects in wide scale range.
Even if enlarging the ISN scale range when integrating FPN, each stage can also be trained well because of the reduced range.
ISN turns to ~\textit{feature pyramid network} (FPN) for help.
Not only does FPN bring more powerful feature representation for objects, but it also enlarges the feasible scale range.

The experiments also verify the positive effect of ISN on FPN. 



\begin{table*}[t]
   \centering
   \begin{tabular}{l|c|cccccc}
      \toprule
         Method            &Backbone      &AP   &AP$_{50}$  &AP$_{75}$  &AP$_{s}$  &AP$_{m}$  &AP$_{l}$ \\
      \midrule
         SNIP (\textit{w} DCN)\cite{singh2018analysis}   &ResNet-50  &43.6 &65.2 &48.8 &26.4 &46.5 &55.8 \\
         SNIPER (\textit{w} DCN)\cite{singh2018sniper}   &ResNet-50  &43.5 &65.0 &48.6 &26.1 &46.3 &56.0 \\
         ISN (\textit{w} DCN) &ResNet-50-FPN  &\textbf{45.3} &\textbf{66.9} &\textbf{50.7} &\textbf{32.1} &\textbf{47.1} &\textbf{54.9} \\
         SNIPER (\textit{w} DCN)\cite{singh2018sniper}   &ResNet-101 &46.1 &67.0 &51.6 &29.6 &48.9 &58.1 \\
         ISN (\textit{w} DCN) &ResNet-101-FPN &\textbf{46.5} &\textbf{67.8} &\textbf{52.0} &\textbf{32.1} &\textbf{48.3} &\textbf{56.4} \\
      \bottomrule
   \end{tabular}
   \vspace{0.1cm}
   \caption{\textbf{SOTA Comparison on COCO test-dev}.
   \textit{w} DCN indicates deformable convolution~\cite{dai2017deformable} is adopted in network.
   All methods are trained on COCO train2017.
   SNIP\cite{singh2018analysis} and SNIPER\cite{singh2018sniper} are recently proposed methods and achieve very impressive performance.
   ISN is integrated with FPN and surpasses previous methods hugely.}
   \label{sota_testdev}
\end{table*}

\begin{table*}[t]
   \centering
   \begin{tabular}{l|c|cccccc}
      \toprule
         Method            &Backbone      &AP   &AP$_{50}$  &AP$_{75}$  &AP$_{s}$  &AP$_{m}$  &AP$_{l}$ \\
      \midrule
         Faster R-CNN\cite{Detectron2018} &ResNet-50-FPN &37.9 &-- & -- &-- &-- & -- \\
      \midrule
         Faster R-CNN (\textit{our impl.})&ResNet-50-FPN &38.0 &58.6 &41.8 &22.0 &41.5 &48.8 \\
         +MS Train\&MS Test               &ResNet-50-FPN &40.9 &62.2 &45.2 &26.9 &43.9 &51.2 \\
         +ISN  &ResNet-50-FPN &\textbf{42.9} &\textbf{64.5} &\textbf{47.5} &\textbf{31.4} &\textbf{45.0} &\textbf{53.9}\\
      \midrule
         Faster R-CNN+DCN                 &ResNet-50-FPN &41.1 &62.1 &45.7 &24.6 &44.2 &53.7\\
         +MS Train\&MS Test               &ResNet-50-FPN &43.0 &64.7 &47.8 &28.0 &46.2 &54.6 \\
         +ISN  &ResNet-50-FPN &\textbf{45.0} &\textbf{66.6} &\textbf{50.3} &\textbf{34.0} &\textbf{48.0} &\textbf{56.3} \\
         SNIPER\cite{singh2018sniper}     &ResNet-101    &46.1 &67.0 &51.6 &29.6 &48.9 &58.1\\
         +ISN  &ResNet-101-FP &\textbf{46.4} &\textbf{67.8} &\textbf{51.9} &\textbf{35.2} &\textbf{49.0} &\textbf{57.7}\\
      \midrule
         SSD\cite{sandler2018mobilenetv2} &MobileNet-v2 &22.1 &-- &-- &-- &-- &-- \\
         SNIPER$^*$\cite{singh2018sniper}          &MobileNet-v2 &34.1 &54.4 &37.7 &18.2 &36.9 &46.2 \\
      \midrule
         Faster R-CNN+ISN                 &MobileNet-v2-FPN &36.6 &57.8 &40.1 &25.5 &38.0 &46.1 \\
         +DCN  &MobileNet-v2-FPN &\textbf{38.7} &\textbf{60.8} &\textbf{42.3} &\textbf{27.8} &\textbf{40.8} &\textbf{48.9} \\
      \bottomrule
   \end{tabular}
   \vspace{0.1cm}
   \caption{\textbf{Comparison on COCO val2017}
   Baseline is trained and tested with single scale (800,1333).
   The MS denotes multi-scale (with 7 scales for training and 9 scales for testing) and the detailed scale setting follows \cite{he2017mask}.
   As is shown, ISN consistently improves APs.
   Methods with superscript * are evaluated on COCO test-dev.}
   \label{sota_val}
\end{table*}

\subsection{ISN on recognition}\label{isnonrecognition}
Inspired by the success of ISN on object detection, we also verify the effect of ISN on other instance-related recognition tasks,
such as instance segmentation and human pose estimation.
Instance segmentation aims to precisely localize and predict pixel-wise mask for all instances in images;
Human pose estimation aims to localize person keypoints of 17 categories accurately.
He et al.,~\cite{he2017mask} proposed an unified framework called Mask R-CNN to solve both tasks.
We try to apply ISN to Mask R-CNN for better recognition accuracy.
In detail, ISN just filters out objects which are out of the given scale range, and use Mask R-CNN in training and testing phases.
\section{Experiments}
\subsection{Common settings}
We conduct experiments on three tasks, namely object detection, instance segmentation, and human pose estimation respectively.
Experiments are performed on COCO~\cite{lin2014microsoft} dataset following the official dataset split.
That is, models are trained on 118k train set and evaluated on the 5k validation set.
The final result of detection for comparison is submitted to official evaluation server of COCO.
\vspace{-0.5cm}
\paragraph{implementation details}\label{common_imple_detail}
If without specific description, the following settings apply to both baseline and ISN.

All models are implemented on the same codebase for comparison based on MXNet\footnote{\href{https://mxnet.apache.org/}{https://mxnet.apache.org/}}.
The training and testing hyper-parameters are almost the same as Mask R-CNN~\cite{he2017mask}, while some modifications were made.

Considering network structures, 2-fc layer detection head exists in all models, which is different from ~\cite{he2016deep} which attaches conv5 as the hidden layers of the head.
For small backbones, such as ResNet-18, ResNet-18-FPN and MobileNet-v2-FPN, the dimension of 2 fc-layers was 512 and 256; while others employed two 1024.
Besides, we use Soft-NMS\cite{bodla2017soft} to replace conventional NMS for post-processing after detection head.
Other settings generally follow Mask R-CNN~\cite{he2017mask}.

For training mechanism, all models were trained on 8 GPUs with synchronized SGD for 18 epochs.
The learning rate ($lr$) is initialized with $0.00125*bs$ which is linear with mini-batch size $bs$ like ~\cite{goyal2017accurate}.
In addition, $lr$ will be divided by $10$ at $15$-th and $18$-th epoch successively. Warm-up strategy~\cite{goyal2017accurate} is also used for the $1$-st epoch.

Baseline models are all trained and tested with $(800, 1333)$ when considering single-scale strategy. 
And multi-scale training \& testing strategy follows ~\cite{he2017mask}.
Concretely, training scales are randomly sampled from $[608, 800]$ pixels with a step of $32$,
while testing scales are $[400, 1200]$ pixels with a step of $100$.
This results in 7 scales for training and 9 scales for testing.

The scaling factor $\Omega$ of ISN models is $\{4.0,2.0,1.0,0.5,0.25\}$ for both training and testing phase.
ISN range is set as $[16, 560]$ by experiments(in Sec.~\ref{exp_ISN_range}).

In the following, experiments of ISN for different recognition tasks are described in details.
\subsection{ISN on object detection}
Detector is based on the two-stage Faster R-CNN~\cite{ren2015faster} framework, i.e. the detection part of Mask R-CNN~\cite{he2017mask}.
Backbones used here consist of vanilla ResNet, ResNet-FPN(ResNet with FPN) and MobileNet-v2-FPN.
We will report standard COCO detection metrics, including AP(mean AP over multiple IoU thresholds),
AP$_{50}$, AP$_{75}$ and AP$_s$, AP$_m$, AP$_l$(AP at different scales).
All training and testing implementation details are the same as ~\ref{common_imple_detail}.
\subsubsection{ISN range}\label{exp_ISN_range}
ISN introduces consistency on scale range, and the only remaining problem is how to find the best ISN range. 
We propose a simple greedy policy that iteratively adjusts the upper bound and lower bound to find the one with the best AP locally.
Experiments on those different range candidates are performed to evaluate AP.
Table~\ref{exp_ISN_scale_range} shows some ISN range candidates with corresponding accuracy statistics.
The following describes the iterative search procedure in detail.

The initial range is set to $[0, 640]$ because of the object size limitation of the COCO dataset.
Firstly $s_u$ is fixed and $s_l\in\{0,16,32\}$ are evaluated respectively.
The AP of $[0, 640]$ is only $37.4$, while AP of $[16, 640]$ increases to $38.2$ because of excluding many hard tiny objects
(about $14\%$ objects' scales in COCO lie in $[0,16]$) during training.
However, further lifting $s_l$ to $32$ deteriorates the AP$_s$ because too many small objects have been ignored in testing phase.
So the locally optimal $s_l$ is $16$.
Secondly $s_l$ is set to $16$ and $s_u\in\{640,560,496,320\}$ are evaluated severally.
From $[16, 640]$ to $[16,560]$, AP, AP$_m$ and AP$_l$ increase because some large objects are filtered during training.
This indicates extremely large objects could disturb learning process doubtlessly.
However, further reducing $s_u$ to $496$ deteriorates AP, in particular AP$_m$, AP$_l$.
We consider that
each stage in FPN learns from objects of different sizes, e.g., the $P_5$
of original FPN receives objects which scale in $[480, \inf]$ by heuristic rule~\cite{lin2017feature}.
And discarding objects in $[496,560]$ causes $P_5$ under-fitting due to lack of training samples.
Analogously, $[16, 320]$ deteriorates AP more. So the locally optimal $s_u$ is $560$.
Thirdly $s_l$ was set to $560$ and $s_u=32$ is evaluated continuously. The AP hasn't been improved.
This iterative search procedure is completed up to now, and the best ISN range is $[16, 560]$.
And ISN range will be set to $[16,560]$ for following experiments.
\begin{table*}[t]
   \centering
   \begin{tabular}{l|ccc|ccc|c}
      \toprule
         ISN\_range   &[0,640] &[16,640] &[32,640] &[16,560] &[16,496] &[16,320] &[32,560]\\ 
      \midrule
         AP$_{s}$    &26.0 &26.9 &25.9 &26.5 &26.2 &26.3 &27.0 \\
         AP$_{m}$    &39.4 &40.1 &40.6 &40.5 &39.8 &39.1 &40.6 \\
         AP$_{l}$    &46.6 &47.6 &47.9 &47.9 &46.7 &46.8 &48.5 \\
         AP          &37.4 &38.2 &38.1 &\textbf{38.7} &37.9 &37.2 &38.4 \\
      \bottomrule
   \end{tabular}
   \vspace{0.1cm}
   \caption{\small{\textbf{ResNet-18-FPN trained with different ISN range}:
   Since object scale in COCO~\cite{lin2014microsoft} ranges from 0 to 640, the trials of range is limited in that region.
   Experimental settings are the same in the table, except the ISN range differs.}}
   \label{exp_ISN_scale_range}
\end{table*}

\begin{table}[b]
   \small
   \centering
   \begin{tabular}{l|l|cccc}
      \toprule
         Method               &Backbone      &AP   &AP$_s$  &AP$_m$  &AP$_l$ \\
      \midrule
         Baseline             &ResNet-18     &29.3 &12.3    &31.2    &42.5 \\
         +MST   &ResNet-18     &31.1 &15.6    &32.6    &44.0 \\
         +ISN                 &ResNet-18     &\textbf{36.1} &\textbf{22.6}  &\textbf{40.5}  &\textbf{47.9} \\
      \midrule
         Baseline             &ResNet-18-FPN &33.3 &17.7    &35.8    &44.0 \\
         +MST   &ResNet-18-FPN &35.6 &21.8    &37.8    &45.7 \\
         +ISN                 &ResNet-18-FPN &\textbf{38.7} &\textbf{26.5} &\textbf{40.5} &\textbf{47.9}\\
      \midrule
         Baseline          &MobileNet-v2-FPN &32.8 &18.5 &35.3 &42.6 \\
         +MST   &MobileNet-v2-FPN &34.3 &21.6 &36.3 &44.3 \\
         +ISN  &MobileNet-v2-FPN &\textbf{36.6} &\textbf{25.5} &\textbf{38.0} &\textbf{46.1} \\
      \bottomrule
   \end{tabular}
   \vspace{0.1cm}
   \caption{\small{\textbf{Results on ResNet-18, ResNet-18-FPN and MobileNet-v2-FPN}:
   Baseline is trained and tested with $[800,1333]$. MST denotes multi-scale training\&testing here.
   Each row adds an individual component to the 1-st row (i.e, Baseline)}}
   \label{res18-frcnn-and-fpn}
\end{table}

\begin{table}[t]
   \centering
   \resizebox{0.45\textwidth}{!}{
      \begin{tabular}{l|cccc}
         \toprule
            MobileNet-v2-FPN  &AP   &AR   &AP$_{[16,560]}$  &AR$_{[16,560]}$\\
         \midrule
            Baseline$_{1.0}$  &27.6 &54.4 &31.0 &60.3 \\
            Baseline$_{800}$  &32.8 &60.8 &35.9 &65.2 \\
         \midrule
            +MS Train$_{1.0}$ &29.1 &55.7 &32.6 &61.4 \\
            +MS Train$_{800}$ &32.9 &61.1 &35.9 &65.4 \\
         \midrule
            +ISN$_{1.0}$      &32.6 &59.4 &\textbf{37.1} &66.9 \\
            +ISN$_{600}$      &33.6 &61.4 &\textbf{37.4} &67.1 \\
         \bottomrule
      \end{tabular}
   }
   \vspace{0.1cm}
   \caption{\textbf{Single scale test on MobileNet-v2-FPN}:
   Baseline is trained in single scale (800,1333).
   The multi-scale trained model and model trained with ISN are shown. Each row is tested with $\omega_i=1.0$ (i.e., raw image)
   or with short side 600 or 800, which is denoted by the subscript. Testing with raw image is much faster than with shorter side 800. $AP_{[16,560]}$ represents the AP calculated on GT boxes in [16,560].
   }
   \label{mobile_sss}
\end{table}
\subsubsection{Main Results}
The comparison of ISN with other state-of-the-art methods on COCO test-dev is shown in Table~\ref{sota_testdev}.
To the best of our knowledge, for ResNet-50-FPN backbone(with deformable convolutions) based on Faster R-CNN architecture,
final mAP on COCO test-dev is $45.3\%$, surpassing previous detectors with same backbone and architecture by a large margin.
ISN, baseline, and other methods are also compared on COCO val-set, as shown in Table~\ref{sota_val}.
With regard to ResNet-50-FPN backbone, AP 38.0 of single scale baseline which is re-implemented is comparable to the one in Detectron~\cite{Detectron2018}.
While multi-scale training \& testing improved baseline by 2.9 point, ISN improved much more to 4.9 point.
Deformable convolution (DCN) was also introduced because of the good property at modeling object deformation.
With DCN integrated, the original baseline goes up to 41.1, and ISN also surpasses the multi-scale competitor again by 2 point gap.
These prove the effectiveness and compatibility of ISN.
In addition, ISN promotes a huge 14.5 point compared the SSDLite~\cite{sandler2018mobilenetv2} version on MobileNet-v2 backbone.

\subsubsection{Results on small backbones}
We also evaluate ISN on other small network architectures, including vanilla ResNet-18, ResNet-18-FPN and MobileNet-v2-FPN. 
With respect to ResNet-18 based Faster R-CNN in Table~\ref{res18-frcnn-and-fpn},
multi-scale training \& testing improves the AP of single scale model by about 2 point,
and ISN significantly boosts incredible 5 point additionally. The AP$_s$, AP$_m$, AP$_l$ get promotion steadily.
Moreover as shown in Table~\ref{res18-frcnn-and-fpn}, using FPN can improve AP of single scale model by 4 point.
Multi-scale FPN further increases AP to $35.6$. ISN with FPN still boosts 3.1 point.
This demonstrates that ISN can bring profit to detector consistently no matter the existence of FPN.
What's more, the vanilla Faster R-CNN with ISN even surpasses the multi-scale FPN without ISN by 0.5 point, manifesting
ISN's superiority.
We also apply ISN to MobileNet-v2-FPN
, as shown in 
Table~\ref{res18-frcnn-and-fpn}.
ISN improves more than 2 point steadily comparing with multi-scale training \& testing.
\subsubsection{ISN for efficient detection}
Although ISN gives model impressive accuracy promotion, it may be criticized owing to its time-consuming multi-scale testing in the real application.
Experiments are conducted to inspect the influence of ISN on tiny models.
And extra comparative experiments are conducted on different single-scale testing circumstances for the same tiny model.
The one is testing on the same pre-defined resolution for all images, while another is testing on the original resolution for each image.
All results are shown in Table~\ref{mobile_sss}.
The trained Baseline, MS train, and ISN model are the same as those in
Table~\ref{res18-frcnn-and-fpn}.
For the 1st single-scale testing circumstance, the pre-defined resolution for Baseline and MS train models is $(800, 1333)$,
while $600$ for ISN because it's closer to the patch size $576$ used in ISN training.
At both testing circumstances, the AP of ISN defeats the other two competitors.
When changing from pre-defined resolution to original resolution, AP of Baseline and MS train declines sharply,
while AP of ISN only decreases 1 point. This also proves ISN's robustness for scale variation.

In real application, there exists a strict requirement of detection accuracy for objects only in common scale range (e.g. $[16, 560]$).
As Table~\ref{mobile_sss} shows, AP$_{[16-560]}$ of ISN surpasses the other two methods by a large margin.
It's because ISN training make model focus on learning object representation in this range.
In summary, ISN achieves better accuracy in a specific scale range while being faster with single-scale testing.
This suggests potential usage of ISN in real situation like security monitoring, etc.
\subsection{ISN on other recognition tasks}
Experiments are also performed on other two object-related recognition tasks, namely instance segmentation and human pose estimation(namely person keypoint detection).
Firstly, Evaluation metrics for both are marked by AP$^{mask}$ and AP$^{kps}$ severally.
The higher AP means more accurate for both, which is similar to detection metric.
Next, for economy, Backbones used here only includes ResNet-FPN (18 and 50), and only multi-scale strategy is compared to ISN.
The following details more.

\begin{table*}[!t]
   \centering
   \resizebox{\textwidth}{!}{
      \begin{tabular}{l|l|c|cccccc|ccc}
         \toprule
            Backbone                         &Method     &AP$^{det}$ &AP$^{mask}$ &AP$^{mask}_{50}$  &AP$^{mask}_{75}$  &AP$^{mask}_{s}$  &AP$^{mask}_{m}$  &AP$^{mask}_{l}$  &AR$^{mask}_{s}$ &AR$^{mask}_{m}$ &AR$^{mask}_{l}$ \\
         \midrule
            \multirow{2}{*}{ResNet-18-FPN}   &MS Train\&MS Test &36.1   &34.1          &53.8          &36.6          &16.2          &36.3  &49.3 &36.6          &56.4          &67.6 \\
                                             &ISN        &\textbf{37.8} &\textbf{35.2} &\textbf{55.4} &\textbf{38.3} &\textbf{20.5} &37.0 &49.6 &\textbf{42.6} &\textbf{58.9} &66.8 \\
         \midrule
            \multirow{2}{*}{ResNet-50-FPN}   &MS Train\&MS Test &41.0   &37.6          &58.3          &40.6          &18.3          &40.8 &53.5 &39.1          &59.5          &70.0\\
                                             &ISN        &\textbf{43.0} &\textbf{39.0} &\textbf{60.0} &\textbf{42.3} &\textbf{22.8} &40.6 &53.3 &\textbf{45.6} &\textbf{61.6} &\textbf{70.6}\\
         \midrule
            \multirow{2}{*}{ResNet-50-FPN-DCN}  &MS Train\&MS Test &42.6   &39.0       &60.2          &42.0          &19.6          &42.1 &56.1 &39.8          &59.6          &70.2\\
                                             & ISN       &\textbf{45.0} &\textbf{40.3} &\textbf{61.9} &\textbf{44.2} &\textbf{24.2} &41.7 &56.8 &\textbf{45.3} &\textbf{61.8} &\textbf{71.3}\\
         \bottomrule
      \end{tabular}
   }
   \vspace{0.1cm}
   \caption{\small{\textbf{Mask Results}: Results of ResNet-18,50 on COCO2017.
   ISN achieves more than 1 point improvement of AP$^{det}$ and AP$^{mask}$ on different backbones.
   Robust and comparable improvements of AR$^{mask}$}}
   \label{tab_mask_backbone_rst_all}
\end{table*}

\begin{table*}[!h]
   \small
      \centering
      \resizebox{0.8\textwidth}{!}{
         \begin{tabular}{l|l|c|ccccc}
            \toprule
               Backbone                         &Method     &AP$^{kps}$ &AP$^{kps}_{50}$ &AP$^{kps}_{75}$ &AP$^{kps}_{m}$ &AP$^{kps}_{l}$\\
            \midrule
               \multirow{3}{*}{ResNet-18-FPN}   &MS Train\&SS Test   &57.8 &81.9 &62.2 &52.7 &65.5 \\
                                                &MS Train\&MS Test   &58.7 &81.7 &63.5 &54.0 &66.6 \\
                                                &\textbf{ISN}        &\textbf{62.5} &\textbf{83.1} &\textbf{68.3} &\textbf{58.2} &\textbf{70.0} \\
               \hline
               \multirow{4}{*}{ResNet-50-FPN}   & Mask R-CNN\cite{he2017mask}  &64.2 &86.6 &69.7 &58.7 &73.0\\
                                                & MS Train\&SS Test  &61.2 &84.1 &66.4 &56.8 &68.2\\
                                                & MS Train\&MS Test  &61.8 &83.7 &66.7 &57.3 &69.5\\
                                                & \textbf{ISN}       &\textbf{65.2} &85.1 &\textbf{71.1} &\textbf{60.8} &72.7\\
               \hline
               \multirow{3}{*}{ResNet-50-FPN-DCN}  & MS Train\&SS Test &62.6 &85.0 &67.9 &57.4 &70.7\\
                                                & MS Train\&MS Test &62.9 &84.4 &68.2 &58.4 &70.7\\
                                                & \textbf{ISN}      &\textbf{66.5} &\textbf{86.0} &\textbf{72.5} &\textbf{61.7} &\textbf{74.4}\\
            \bottomrule
         \end{tabular}
      }
      \vspace{0.1cm}
      \caption{\small{\textbf{ISN v.s. MST}: AP of person instance detection and keypoint detection on COCO val2017.
      The backbone is ResNet-18-FPN, ResNet-50-FPN without and with DCN. ISN improves more than 3.5 point on AP$^{kps}$ comparing with MST for both backbones.
      The accuracy promotion on kps is steady whether with DCN or not, proving ISN is compatible with DCN.
      For comparison to official Mask R-CNN, the MS Train\&MS Test with modified kps head is quicker and bandwidth-saving in practice, but inferior on accuracy.
      The ISN model with the same modified kps head can defeat Mask R-CNN\cite{he2017mask} easily.
      }}
      \label{kps_baseline}
   \end{table*}
\subsubsection{ISN on instance segmentation}
\paragraph{Main results}
Results of detection and instance segmentation on different backbones are shown in Table~\ref{tab_mask_backbone_rst_all}.
As you can see, AP on mask for ISN achieves more than 1 point increment comparing with multi-scale training \& testing in both shallow and deep backbones.
Even if introducing DCN component, the increment is also obvious and robust. This indicates the compatibility of ISN and other strong components (e.g. DCN).
It's worth noting that ISN can also improve AR metric on objects of different sizes. This is essential to real applications.
\subsubsection{ISN on human pose estimation}
Generally speaking, there are always two modes of human pose estimation: 1) From image, namely detecting person bounding-box,
then cropping it from original image and resizing to standard scale, finally using backbone predicting keypoint heatmaps directly.
2) From feature map, namely detecting person instance and corresponding keypoints simultaneously, which is equivalent to predicting heatmaps from feature map at inter layer.
The former mode is always more accurate than the latter, because of the explicit normalization operation on scale in image, i.e., cropping and resizing to uniform resolution.
Experiments on ISN for kps are performed on COCO-kps data and Mask R-CNN structure, namely the 2nd mode.
The final accuracy boosting on pose estimation in results implies the effect of implicit normalization from ISN.
\paragraph{Implementation details}\label{ISN_kps_detail}
Original Mask R-CNN~\cite{he2017mask} models a keypoint's location distribution on RoI as an one-hot mask,
and predicts $K=17$ heatmaps, each for one keypoint type.
However, the kps head (the head for human pose estimation) is too heavy for real application, especially for multi-person pose estimation.
Therefore, Four modifications on head's structure are made for both ISN and baseline model.
Firstly, to better approximate person's shape, we use RoIAlign to extract feature map of $22\times16$ size for each person RoI, instead of $14\times14$.
Secondly, to reduce computation complexity and bandwidth consumption, the channel number of the eight $3\times3$ conv layers in kps head is reduced from 512 to 256,
and the last deconv layer is also removed.
Thirdly, for each keypoint category, kps head will output corresponding dense heatmap to predict whether each position is in vicinity of current keypoint.
Finally, the head will also output corresponding 2-D local offset vector for each category and position as ~\cite{papandreou2017towards}, to get keypoint localization more precise.
But different from ~\cite{papandreou2017towards}, smooth-L1 loss was used to optimize both outputs severally.
In addition, patches used in ISN training here are generated from COCO keypoint data, which includes
bounding-box and keypoint information.


\paragraph{Main results}
Results of person detection and pose estimation on different backbones are shown in Table~\ref{kps_baseline}.
Firstly, official Mask R-CNN~\cite{he2017mask} is compared to ISN and MS train models on ResNet-50-FPN backbone in the 2-nd row in Table~\ref{kps_baseline}.
It should be pointed out that, the modified design reduces nearly half float operation number for kps head.
We observe that the MS train model is inferior on AP$^{kps}$ to official Mask R-CNN~\cite{he2017mask}.
Because of the relative change of head capacity from reduced kps head to original detection head, 
the training process might pay more attention to detection branch. This contrary effect is also shown in ~\cite{he2017mask}.
However, even with the same light-weight design, ISN still exceeds Mask R-CNN~\cite{he2017mask} by 1 point on AP$^{kps}$.
This proves the effectiveness of ISN for human pose estimation.

Secondly, ISN boosts performance of human pose estimation on several backbones significantly, as shown in Table~\ref{kps_baseline}.
The AP$^{kps}$ gets more close to kps method~\cite{xiao2018simple} from image, namely the aforementioned 1st mode, than Mask R-CNN~\cite{he2017mask}.
This phenomenon indicates that implicit scale normalization in ISN can assist kps, consistently with
the explicit scale normalization in 1st mode method.

\section{Conclusion}
We propose a novel paradigm, instance scale normalization to solve severe scale variation problem in instance-related vision tasks.
ISN integrates image pyramid (for scale normalization) and feature pyramid (for easeful learning in ISN range)
in one paradigm, and achieves enhanced scale processing capability.
It significantly boosts the performance of multi-scale training \& testing on object recognition tasks,
i.e., object detection, instance segmentation, and human pose estimation over strong baselines.
It also can be extended to more efficient detection for real applications.
Overall, ISN provides a new perspective to solve problem raising from scale.
It should inspire future work to solve other large variation problems.
{\small
\bibliographystyle{ieee}
\bibliography{ISN_detector}
}

\end{document}